\documentclass[]{article}
\usepackage{proceed2e-onecolumn}

\usepackage{times}

\usepackage{listings}
\usepackage{tikz}
\usetikzlibrary{fit,positioning}

\usepackage{comment}
\usepackage{bm}
\usepackage[parfill]{parskip}

\usepackage[authoryear,round]{natbib}
\usepackage{amsmath}
\usepackage{amsthm}
\usepackage{amssymb}

\usepackage{amsfonts}
\graphicspath{{figures/}}

\newcommand{\pushleft}[1]{\ifmeasuring@#1\else\omit$\displaystyle#1$\hfill\fi\ignorespaces}

\newcommand{\R}{{\mathbb{R}}}
\newcommand{\p}{{\mathsf{P}}}
\newcommand{\q}{{\mathsf{Q}}}

\newcommand{\dataset}{{\{x_i\}_{i=1}^n\sim \p}}
\newcommand{\observations}{{\{x_i\}_{i=1}^n}}
\newcommand{\embeddingp}{{\mu_{\p}}}
\newcommand{\embeddingq}{{\mu_{\q}}}

\newcommand{\domX}{{\mathcal{X}}}
\newcommand{\domY}{{\mathcal{Y}}}
\newcommand{\rkhs}{{\mathcal{H}_k}}
\newcommand{\sfS}{{\mathsf{S}}}

\newcommand\ci{\protect\mathpalette{\protect\cI}{\perp}}
\def\cI#1#2{\mathrel{\rlap{$#1\;#2$}\mkern2mu{#1#2\;}}}

\newtheorem{theorem}{Theorem}
\newtheorem{definition}{Definition}

\title{Bayesian Learning of Kernel Embeddings}

\author{ {\bf Seth Flaxman} \\
flaxman@stats.ox.ac.uk \\
Department of Statistics \\
University of Oxford
\And
{\bf Dino Sejdinovic}  \\
dino.sejdinovic@stats.ox.ac.uk \\
Department of Statistics \\
University of Oxford
\And
{\bf John P. Cunningham}   \\
jpc2181@columbia.edu \\
Department of Statistics \\
Columbia University 
\And
{\bf Sarah Filippi}   \\
filippi@stats.ox.ac.uk \\
Department of Statistics \\
University of Oxford
}

\begin{document}

\maketitle
\begin{abstract} 
Kernel methods are one of the mainstays of machine learning, but the problem of
kernel learning 
remains challenging, with only a few heuristics and very
little theory. This is of particular importance in methods based
on estimation of kernel mean embeddings of probability measures. For characteristic kernels, which include most commonly used ones, the kernel mean embedding uniquely determines its probability measure, so it can be used
to design a powerful statistical testing framework, which includes nonparametric two-sample and independence tests. In practice,
however, the performance of these tests can be very sensitive to the choice of kernel
and its lengthscale parameters. To address this central issue, we propose a new probabilistic model for 
kernel mean embeddings, the Bayesian Kernel Embedding model, combining a Gaussian process prior over 
the Reproducing Kernel Hilbert Space containing the mean embedding with a conjugate likelihood function, thus yielding a closed form
posterior over the mean embedding. The posterior mean of our model is closely related to recently
proposed shrinkage estimators for kernel mean embeddings, while the posterior uncertainty is a new, interesting feature 
with various possible applications.
Critically for the purposes of
kernel learning, our model gives a simple, closed form
marginal pseudolikelihood of the observed data given the kernel hyperparameters. This marginal pseudolikelihood can either be optimized to inform the hyperparameter choice or fully Bayesian inference can be used.
\end{abstract}

\section{INTRODUCTION}
\label{introduction}
A large class of popular and successful machine learning methods rely on
kernels (positive semidefinite functions), including support
vector machines, kernel ridge regression, kernel PCA \citep{scholkopf2002learning}, Gaussian processes \citep{rasmussen2006gaussian}, and kernel-based hypothesis testing \citep{gretton2005measuring,gretton2008kernel,gretton2012kernel}.
A key component for many of these methods is that of estimating kernel mean embeddings and covariance operators of probability measures based on data. The use of simple empirical estimators has been challenged recently \citep{muandet2015kernel} and alternative, better-behaved frequentist shrinkage strategies have been proposed. 
In this article, we develop a Bayesian framework for estimation of kernel mean embeddings, recovering desirable shrinkage properties as well as allowing quantification of full posterior uncertainty. Moreover, the developed framework has an additional extremely useful feature. Namely, a persistent problem in kernel methods is that of kernel choice and hyperparameter selection, for which no general-purpose strategy exists. When a large dataset is available in a supervised
setting, the standard approach is to 
use cross-validation. However, in unsupervised learning and kernel-based hypothesis testing, cross-validation is not straightforward to apply and yet the choice of kernel is critically important. Our framework gives a tractable closed-form marginal pseudolikelihood of the data allowing direct hyperparameter optimization as well as fully Bayesian posterior inference through integrating over the kernel hyperparameters. We emphasise that this approach is fully unsupervised: it is based solely on the modelling of kernel mean embeddings -- going beyond marginal likelihood based approaches in, e.g., Gaussian process regression -- and is thus broadly applicable in situations, such as kernel-based hypothesis testing, where the hyperparameter choice has thus far been mainly driven by heuristics.
  
In Section \ref{section:background} we provide the necessary background on Reproducing Kernel Hilbert Spaces (RKHS) as well as describe some related works.
In Section \ref{section:bke} we develop our Bayesian Kernel Embedding model, showing a rigorous Gaussian
process prior formulation for an RKHS. In Section \ref{section:bkl} we show how to perform kernel learning and posterior inference with our model. In Section \ref{section:experiments} we empirically evaluate our model,
arguing that our Bayesian Kernel Learning (BKL) objective should be considered as a ``drop-in'' replacement for heuristic methods of choosing kernel hyperparameters currently in use, especially in unsupervised
settings such as kernel-based testing. We close in Section \ref{section:discussion} with a discussion of various applications of our approach and future work.

\section{BACKGROUND AND RELATED WORK}
\label{section:background}
\subsection{KERNEL EMBEDDINGS OF PROBABILITY MEASURES}
For any positive definite kernel function $k:\domX\times\domX\to\R$, there exists a unique reproducing kernel Hilbert space (RKHS) $\rkhs$.  RKHS is an (often infinite-dimensional) space of functions $h:\domX\to\R$ where evaluation can be written as an inner product, and in particular $h(x)=\langle h,k(\cdot,x) \rangle_\rkhs$ for all $h\in\rkhs, x\in\domX$. Given a probability measure $\p$ on $\domX$, its kernel
embedding into $\rkhs$ is defined as:
\begin{equation}
   \embeddingp =\int k\left(\cdot,x\right) \p(dx).
 \label{eq:kme2}
\end{equation}
Embedding $\embeddingp$ is an element of $\rkhs$ and serves as a representation of $\p$ akin to a characteristic function. It represents expectations of RKHS functions in the form of an inner product $\int h(x)\p(dx)=\langle h,\embeddingp \rangle_\rkhs$. 
For a broad family of kernels termed \emph{characteristic} \citep{sriperumbudur2011}, every probability measure has a unique embedding -- thus, such embeddings completely determine their probability measures and capture all of the moment information. This yields a framework for constructing nonparametric hypothesis tests for the two-sample problem and for independence, which are consistent against all alternatives \citep{gretton2008kernel,gretton2012kernel} -- we review this framework in the next section.

\subsection{KERNEL MEAN EMBEDDING AND HYPOTHESIS TESTING}
\label{section:mmd-hsic-kpca}
Given a kernel $k$ and probability measures $\p$ and $\q$, the maximum mean discrepancy (MMD) between $\p$ and $\q$ \citep{gretton2012kernel} is defined as the squared RKHS distance $\Vert \embeddingp-\embeddingq \Vert^2_\rkhs$ between their embeddings.
A related quantity is the Hilbert Schmidt Independence Criterion (HSIC) \citep{gretton2005measuring,gretton2008kernel}, a nonparametric dependence measure between random variables $X$ and $Y$ on domains $\domX$ and $\domY$ respectively, defined as the squared RKHS distance $\Vert \mu_{\p_{XY}}-\mu_{\p_X\p_Y} \Vert^2_{\mathcal{H}_\kappa}$ between the embeddings of the joint distribution $\p_{XY}$ and of the product of the marginals $\p_X\p_Y$ with respect to a kernel $\kappa:(\domX\times\domY)\times(\domX\times\domY)\to\R$ on the product space. Typically, $\kappa$ factorises, i.e. $\kappa\left((x,y),(x',y')\right)=k(x,x')l(y,y')$. The empirical versions of MMD and HSIC are used as test statistics for the two-sample (${\bf H}_0: \p = \q$ vs. ${\bf H}_1: \p \neq \q$) and independence (${\bf H}_0: X \ci Y$ vs. ${\bf H}_1: X \not\ci Y$) tests, respectively. With the help of the approximations to the asymptotic distribution under the null hypothesis, corresponding p-values can be computed \citep{gretton2012kernel}. In addition, the so-called ``witness function'' which is proportional to $\embeddingp - \embeddingq$ can be used to assess where the difference between the distributions arises.

\subsection{KERNEL MEAN EMBEDDING ESTIMATORS}
\label{section:kme:estimators}
For a set of i.i.d. samples $x_1, \ldots, x_n$, the kernel mean embedding is typically estimated by its empirical version
\begin{equation}
    \widehat{\embeddingp} = \mu_{\widehat\p}= \frac{1}{n}\sum_{i=1}^n k(\cdot,x_i),
\label{eq:kme-estimator}
\end{equation}
from which various associated quantities, including the estimators of the squared RKHS distances between embeddings needed for kernel-based hypothesis tests, follow. As an empirical mean in an infinite-dimensional space, \eqref{eq:kme-estimator} is affected by Stein's phenomenon, as overviewed by  \citet{muandet2013kernel} who also propose alternative shrinkage estimators similar to the well known James-Stein estimator. Improvements of test power using such shrinkage estimators are reported by \citet{ramdaswehbe2015}. Connections between the James-Stein estimator and empirical Bayes procedures are classical \citep{efron1973stein}, and thus a natural question to consider is whether a Bayesian formulation of the problem of kernel embedding estimation would yield similar shrinkage properties. In this paper, we will give a Bayesian perspective of the problem of kernel embedding estimation. In particular, we will construct a flexible model for underlying probability measures based on Gaussian measures in RKHSs which allows derivation of a full posterior distribution of $\embeddingp$, recovering similar shrinkage properties to \citet{muandet2013kernel}, as discussed in Section \ref{section:relation_shrinkage}. The model will give us a further advantage, however -- as the marginal likelihood of the data given the kernel parameter can be derived leading to an informed choice of kernel parameters.

\subsection{SELECTION OF KERNEL PARAMETERS }
\label{section:selection:parameterer}

In supervised kernel methods like support vector machines, leave-one-out or k-fold crossvalidation is an effective and
widely used method for kernel selection, and the myriad papers on multiple kernel learning (e.g.~\cite{bach2004multiple,sonnenburg2006large,gonen2011multiple}) assume
that some loss function is available and thus focus on effective ways of learning combinations of kernels.
In the related but distinct world of smoothing kernels and kernel density estimation, there are a variety of
long-standing approaches to bandwidth selection, again based on a loss function (in this case, mean integrated squared error is a popular choice \citep{bowman1985comparative}, and there is even a formula giving the optimal smoothing parameter asymptotically, see \cite{rosenblatt1956remarks,parzen1962estimation}) but we are not aware of work linking this literature to methods based on positive definite/RKHS kernels we study here.
Separately, Gaussian process learning can be undertaken by maximizing the marginal likelihood, which has a convenient closed form. This is noteworthy for its success and general applicability even for learning complicated combinations of kernels \citep{duvenaud2013structure} or rich kernel families \citep{wilson2013gaussian}. Our approach has the same basic design as that of Gaussian process learning, yet it is applicable to learning kernel embeddings, which falls outside the realm of  supervised learning.

As noted in \citet{gretton2012optimal}, the choice of the kernel $k$ is critically important for the power of the tests presented in Section~\ref{section:mmd-hsic-kpca}. However, no general, theoretically-grounded approaches for kernel selection in this context exist. The difficulty is that, unlike in supervised kernel methods, a simple cross-validation approach for the kernel parameter selection is not possible. What would be an ideal objective function -- asymptotic test power -- cannot be computed due to a complicated asymptotic null distribution. Moreover, even if we were able to estimate the power by performing tests on ``training data'' for each of the individual candidate kernels, in order to account for multiple comparisons, this training data would have to be disjoint from the one on which the hypothesis test is performed, which is clearly wasteful of power and appropriate only in the type of large-scale settings discussed in \cite{gretton2012optimal}. For these reasons, most users of kernel hypothesis tests in practice resort to using a parameterized kernel family such as squared exponential, and setting the lengthscale parameter based on the ``median heuristic.'' 

The exact origins of the median heuristic  are unclear (interestingly, it does not appear in the book that is most commonly cited as its source, \cite{scholkopf2002learning}) but it may have been derived from \cite{takeuchi2006nonparametric} and has precursors in classical work on bandwidth selection for kernel density estimation \citep{bowman1985comparative}. Note that there are two versions of the median heuristic in the literature: in both versions, given a set of observations $x_1, \ldots, x_n$ we calculate $\ell = \mbox{median}(\|x_i-x_j\|_2)$ and then one version (e.g.~\cite{mooij2015distinguishing}) uses the
Gaussian RBF / squared exponential kernel parameterized as $k(x,x') = \mbox{exp}(-\frac{\|x-x'\|^2}{\ell^{2}})$ and the second
version (e.g.~\cite{muandet2014kernel}) uses the parameterization $k(x,x') = \mbox{exp}(-\frac{\|x-x'\|^2}{2\ell^{2}})$. 
Some recent work has highlighted the situations in which the median heuristic can lead to poor performance \citep{gretton2012optimal}. Cases in which the median heuristic performs quite well and also cases in which it performs quite poorly are discussed in \citep{reddi2015high,ramdas2015decreasing}. We note that the median heuristic has also been used as a default value for supervised learning tasks (e.g.~for the SVM implementation in R package \texttt{kernlab}) or when cross-validation is simply too expensive.  

Outside of kernel methods, the same basic conundrum arises in spectral clustering in the choice of the parameters for the similarity graph \citep[Section 8.1]{vonluxburg2007tutorial} and it is implicitly an issue in any unsupervised statistical method
based on distances or dissimilarities, like the distance covariance (which is in fact equivalent to HSIC with a certain family of kernel functions \citep{sejdinovic2013equivalence}), or even the choice of the number of neighbors $k$ in $k$-nearest neighbors algorithms.

\section{OUR MODEL: BAYESIAN KERNEL EMBEDDING}
\label{section:bke}
Below, we will work with a parametric family of kernels $\{k_\theta(\cdot,\cdot)\}_{\theta\in\Theta}$. Given a dataset $\dataset$ of observations in $\mathbb R^D$ for an unknown probability distribution
$\p$, we wish to infer the kernel embedding $\mu_{\p,\theta}=\int k_\theta\left(\cdot,x\right) \p(dx)$ for a given kernel $k_\theta$ in the parametric family. Moreover, we wish to construct a model that will allow inference of the kernel hyperparameter $\theta$ as well. Note that the two goals are related, since $\theta$ determines the space in which the embedding $\mu_{\p,\theta}$ lies. When it is obvious from context, we suppress the dependence of the embeddings on the underlying measure $\p$, writing $\mu_{\theta}$ to emphasize the dependence on $\theta$. Similarly, we will use $\widehat{\mu_\theta}$ to denote the simple empirical estimator from Eq.~\eqref{eq:kme-estimator}, which depends on a fixed sample $\observations$.

Our Bayesian Kernel Embedding (BKE) approach consists in specifying a prior on the kernel mean embedding $\mu_\theta$ and a likelihood function linking it to the observations through the empirical estimator $\widehat{\mu_\theta}$.  This will then allow us to infer the posterior distribution of the kernel mean embedding. The hyperparameter $\theta$ can itself have a prior, with the goal of learning a posterior distribution over the hyperparameter space.
\subsection{PRIOR}
A given hyperparameter $\theta$ (which can itself have a prior distribution),
parameterizes a kernel $k_{\theta}$ and a corresponding RKHS $\mathcal{H}_{k_{\theta}}$. 
While it is tempting to define a $\mathcal{GP}(0, k_\theta(\cdot,\cdot))$ prior on $\mu_\theta$, this is problematic since draws from such prior would almost surely fall outside $\rkhs$ \citep{wahba1990}. Therefore, we define a GP prior over  $\mu_\theta$ as follows:
\begin{align}
\mu_\theta ~|~ \theta & \sim \mathcal{GP}(0, r_\theta(\cdot,\cdot))\;, \label{eq:prior}\\
    r_{\theta}(x,y) &:= \int k_\theta(x,u)k_\theta(u,y) \nu(du)\;. \label{eq:convolution}
\end{align}
where $\nu$ is any finite measure on $\mathcal X$. This choice of $r_{\theta}$ ensures that $\mu_\theta \in \mathcal{H}_{k_{\theta}}$ with probability 1 by the \emph{nuclear dominance} \citep{Lukic:2001,pillai2007characterizing} of $k_{\theta}$ over $r_{\theta}$ for any stationary kernel $k_\theta$ and more broadly whenever $\int k_\theta(x,x)\nu(dx)<\infty$. For completeness, we provide details of this construction in the Appendix in Section \ref{section:nuclear-dominance}.  
Since Eq.~\eqref{eq:convolution} is the convolution of a kernel
with itself with respect to $\nu$, for typical kernels $k_{\theta}$, the resulting kernel $r_{\theta}$ can be thought of as a smoother
version of $k_{\theta}$. A particularly convenient choice for $\mathcal X=\mathbb R^D$ is to take $\nu$ to be proportional to a Gaussian measure in which case $r_\theta$ can be computed analytically for a squared exponential kernel $k_\theta$. The derivation is given in the Appendix in Section \ref{section:r-calculation}, where we further show that if we set $\nu$ to be proportional to an isotropic Gaussian measure with a large variance parameter,  $r_{\theta}$ becomes very similar to a squared exponential kernel with lengthscale $\theta \sqrt{2}$.

\subsection{LIKELIHOOD}
We need a likelihood linking the kernel mean embedding $\mu_\theta$ to the observations $\observations$.
We define the likelihood via the empirical mean embedding estimator of Eq.~\eqref{eq:kme-estimator},
 $\widehat{\mu_{\theta}}$ which depends on $\observations$ and $\theta$.
Consider evaluating $\widehat{\mu_{\theta}}$ at some $x\in\R^D$ (which need not be one of our observations).
The result is a real number giving an empirical estimate of $\mu_{\theta}(x)$ based on $\observations$ and $\theta$. 
We link the empirical estimate, $\widehat{\mu_{\theta}}(x)$, to the corresponding modeled estimate,
$\mu_{\theta}(x)$ using a Gaussian distribution with variance $\tau^2/n$:
\begin{equation}
p(\widehat{\mu_\theta}(x)|\mu_{\theta}(x)) = \mathcal{N}(\widehat{\mu_\theta}(x); \mu_{\theta}(x), \tau^2/n),\quad x\in\domX.
\label{eq:like1}
\end{equation}
Our motivation for choosing this likelihood comes from the Central Limit Theorem. For a  fixed location $x$, $\widehat{\mu_\theta}(x)=\frac{1}{n}\sum_{i=1}^n k_\theta(x_i,x)$ is an average of i.i.d. random variables so it satisfies:
\begin{equation}
    \sqrt n(\widehat{\mu_\theta}(x) - \mu_\theta(x)) \overset{D}{\rightarrow} \mathcal{N}(0,\text{Var}_{X \sim \p}[k_\theta(X,x)]).
\end{equation}
We note that considering a heteroscedastic variance dependent on $x$ in \eqref{eq:like1} would be a straightforward extension to our model, but we do not pursue this idea further here, i.e. while $\tau^2$ can depend both on $\theta$ and $x$, we treat it as a single hyperparameter in the model. 
 
\subsection{JUSTIFICATION FOR THE MODEL}
There are various ways to understand the construction of our hierarchical model. 
$\observations$ are drawn iid from $\p$, which we do not have access to. We
could estimate $\p$ directly (e.g.~with a Gaussian mixture model) obtaining $\hat\p$, and then estimate $\mu_{\theta,\hat\p}$. But since density estimation is challenging in high dimensions, we posit a generative 
model for $\mu_{\theta}$ directly.

\begin{figure}[h!]
   \centering
    \includegraphics{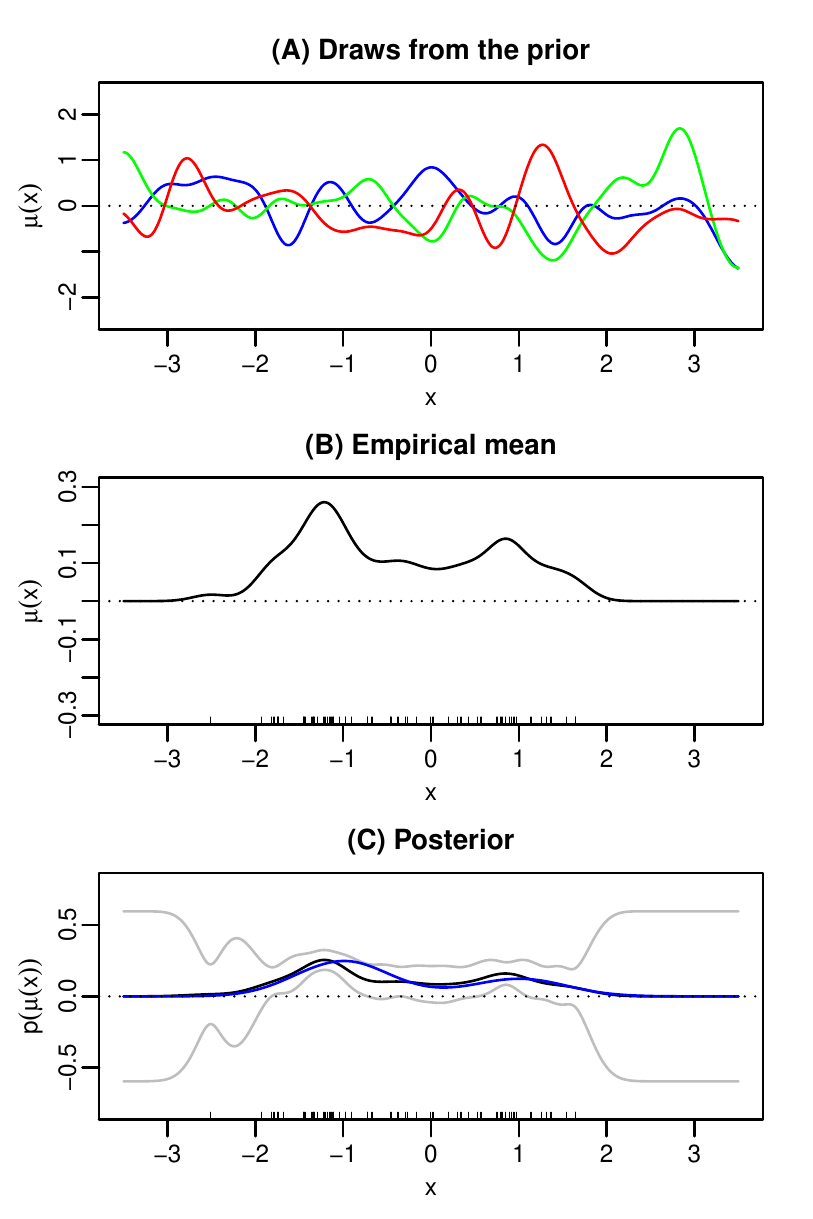}
    \caption{An illustration of the Bayesian Kernel Embedding model, where $k_{\theta}$ is a squared
exponential kernel with lengthscale 0.1. Three draws of $\mu_{\theta}$ from the prior are shown in (A). The
empirical mean estimator $\widehat{\mu_{\theta}}$, which is the link function for the likelihood, is shown in (B) with the observations shown as a rug plot. In (C), the posterior mean embedding (black line) with uncertainty intervals (gray lines) is shown, as is the true mean embedding (blue line) based on the true data generating process (a mixture of Gaussians) and the same $k_{\theta}$. }
    \label{fig:prior-and-likelihood-illustrations}
\end{figure}

 Beginning at the top of the hierarchy, we have a fixed or random hyperparameter $\theta$,
which immediately defines $k_{\theta}$ and the corresponding RKHS $\mathcal{H}_{k_{\theta}}$. Then, we
introduce a GP prior over $\mu_{\theta}$ to ensure that $\mu_{\theta} \in \mathcal{H}_{k_{\theta}}$. 
A few realizations of $\mu_{\theta}$ drawn from our prior are shown in 
Figure \ref{fig:prior-and-likelihood-illustrations} (A), for an illustrative one-dimensional example where the prior is a Gaussian process with squared exponential kernel
with lengthscale $\theta = 0.25$. Small values of $\theta$ yield rough functions and large values of $\theta$ yield smooth functions.
Next, we need to define the likelihood, which links these draws from the
prior to the observations $\observations$. Since $\mu_{\theta}$ is an infinite dimensional
element in a Hilbert space and $\observations \in \mathcal{X}$ we need to transform the observations
so that we can put a probability distribution over them. We use the empirical estimate
of the mean embedding $\widehat{\mu_{\theta}}$ as our link function. Given a few observations,
$\widehat{\mu_{\theta}}$ is shown in Figure \ref{fig:prior-and-likelihood-illustrations} (B). Our likelihood
links $\widehat{\mu_{\theta}}$ to $\mu_{\theta}$ at the observation locations $\observations$ by
assuming a squared loss function, i.e.~Gaussian errors. As mentioned above, the motivation
is the Central Limit Theorem, but also the convenient conjugate form that a Gaussian process with
Gaussian likelihood yields. A plot of the posterior over the mean embedding is shown in Figure \ref{fig:prior-and-likelihood-illustrations} (C). A few points are worth noting: since the empirical estimator is already quite smooth (notice
its similarity to a kernel density estimate), the posterior mean embedding is only slightly smoother than
the empirical mean embedding. Notice that unlike kernel density estimation, there is no requirement that
the kernel mean embedding be non-negative, thus explaining the posterior uncertainty intervals which are below zero.

Our original motivation for considering a Bayesian model for kernel mean embeddings was to see whether there
was a coherent Bayesian formulation that corresponded to the shrinkage estimators in \cite{muandet2013kernel},
while also enabling us to learn the hyperparameters. The first difficulty we faced was how to define
a valid prior over the RKHS and a reasonable likelihood function. Our choices are by no means definitive,
and we hope to see further development in this area in the future. The second difficulty was that of
developing a method for inferring hyperparameters, to which we turn in the next section.

\section{BAYESIAN KERNEL LEARNING}
\label{section:bkl}
In this section we show how to perform learning and inference in the Bayesian Kernel Embedding
model introduced in the previous section. Our model inherits various attractive properties from the
Gaussian process framework \citep{rasmussen2006gaussian}.
First, we derive the posterior and posterior predictive distributions for the kernel mean embedding in closed form due to the conjugacy of our model, and show the relationship with previously proposed shrinkage estimators. We then derive  the tractable marginal likelihood of the observations given the hyperparameters allowing for efficient MAP estimation or posterior inference for hyperparameters. 

\subsection{POSTERIOR AND POSTERIOR PREDICTIVE DISTRIBUTIONS}
\label{section:posterior}
Similarly to GP models, the posterior mean of $\mu_\theta$ is
available in closed form due to the conjugacy of Gaussians.
Perhaps given our data we wish to infer $\mu_\theta$ at a new location $x^*\in\R^D$.
Given a value of the hyperparameter $\theta$ 
we can calculate the posterior distribution of $\mu_\theta$ as well as the posterior predictive distribution $p(\mu_\theta(x^*)|\widehat{\mu_\theta},\theta)$. 

Standard GP results \citep{rasmussen2006gaussian} yield the posterior distribution as:
\begin{align}
   & [\mu_\theta(x_1), \ldots, \mu_\theta(x_n)]^{\top} \;| \;[\widehat{\mu_\theta}(x_1), \ldots, \widehat{\mu_\theta}(x_n)]^\top, \theta \nonumber\\
   & \qquad\sim \mathcal{N}(R_\theta ( R_\theta+(\tau^2/n) I_n)^{-1}[\widehat{\mu_\theta}(x_1), \ldots, \widehat{\mu_\theta}(x_n)]^{\top},\nonumber \\
   &\qquad \qquad \qquad \qquad R_\theta - R_\theta(R_\theta+(\tau^2/n) I_n)^{-1}R_\theta),
\label{eq:posterior-dist}
\end{align}
where $R_\theta$ is the $n\times n$ matrix such that its $(i,j)$-th element is $r_\theta(x_i,x_j)$.
The posterior predictive distribution at a new location $x^*$ is:
\begin{align}
   & \mu_\theta(x^*)^{\top}\; |\; [\widehat{\mu_\theta}(x_1), \ldots, \widehat{\mu_\theta}(x_n)]^\top, \theta \nonumber\\
    & \qquad \sim \mathcal{N}(R_\theta^{*\top} ( R_\theta+(\tau^2/n) I_n)^{-1}[\widehat{\mu_\theta}(x_1), \ldots, \widehat{\mu_\theta}(x_n)]^{\top},\nonumber\\
    & \qquad \qquad \qquad \qquad r_\theta^{**} - R_\theta^{*\top}(R_\theta+(\tau^2/n) I_n)^{-1}R_\theta^*)
    \label{eq:posterior-f}
\end{align}
where $R_\theta^* = \left[ r_\theta(x^*,x_1),  \dots  r_\theta(x^*,x_n)\right]^\top$
    and $r_\theta^{**}=r_\theta(x^*,x^*)$.

As in standard GP inference, the time complexity is $\mathcal{O}(n^3)$ due to the matrix inverses
and the storage is $\mathcal{O}(n^2)$ to store the $n \times n$ matrix $R_{\theta}$.

\subsection{RELATION TO THE SHRINKAGE ESTIMATOR}
\label{section:relation_shrinkage}
The spectral kernel mean shrinkage estimator (S-KMSE) of \cite{muandet2013kernel} for a fixed kernel $k$ is defined as:
\begin{equation}
    \check\mu_{\lambda} = \hat\Sigma_{XX}(\hat\Sigma_{XX}+\lambda I)^{-1} \hat\mu,
    \label{eq:skmse}
\end{equation}
where $\hat\mu=\sum_{i=1}^n k(\cdot,x_i)$ is the empirical embedding, $\hat\Sigma_{XX}=\frac{1}{n}\sum_{i=1}^n k(\cdot,x_i)\otimes k(\cdot,x_i)$ is the empirical covariance operator on $\rkhs$, and $\lambda$ is a regularization parameter. \cite[Proposition 12]{muandet2013kernel} shows that $\check\mu_{\lambda}$ can be expressed as a weighted kernel mean $\check\mu_\lambda=\sum_{i=1}^n \beta_i k(\cdot,x_i)$, where
\begin{eqnarray*}
 \beta&=&\frac{1}{n}(K+n\lambda I)^{-1}K{\bf 1}\\
 {}&=&(K+n\lambda I)^{-1}[\widehat{\mu}(x_1), \ldots, \widehat{\mu}(x_n)]^\top.
\end{eqnarray*}
Now, evaluating S-KMSE at any point $x^*$ gives
\begin{eqnarray*}
 \check{\mu}_{\lambda}(x^*)&=&\sum_{i=1}^n \beta_i k(x^*,x_i)\\
    {}&=&K_{*}^{\top}(K+n\lambda I)^{-1}[\widehat{\mu}(x_1), \ldots, \widehat{\mu}(x_n)]^\top,
\end{eqnarray*}
where $K_{*}=\left[ k(x^*,x_1),  \ldots,  k(x^*,x_n)\right]^\top$.
Thus, the posterior mean in Eq.~\eqref{eq:posterior-dist} recovers the S-KMSE estimator \citep{muandet2013kernel}, where the regularization parameter is related to the variance in the likelihood model \eqref{eq:like1}, with a difference that in our case the kernel $k_\theta$ used to compute the empirical embedding is not the same as the kernel $r_\theta$ used to compute the kernel matrices.
We note that our method has various advantages over the frequentist estimator $\check\mu_{\lambda}$: we have a closed-form uncertainty estimate, while we are not aware of a principled way of calculating the standard error of the frequentist estimators of embeddings. Our model also leads to a method for learning the hyperparameters, which we discuss next. 

\subsection{INFERENCE OF THE KERNEL PARAMETERS}
\label{eq:marg-lik}
In this section we focus on hyperparameter learning in our model.  
For the purposes of hyperparameter learning, we want to integrate
out the kernel mean embedding $\mu_\theta$ and consider the probability of our observations $\observations$ given the hyperparameters $\theta$. 
In order to link our generative model directly to the observations, we use a pseudolikelihood approach as discussed in detail below.

We use the term pseudolikelihood because the model in this section will not correspond to the likelihood of the infinite dimensional empirical embedding; rather it will rely on the evaluations of the empirical embedding at a finite set of points.  Let us fix a set of points $z_{1},\ldots,z_{m}$ in $\mathcal{X}\subset \mathbb R^D$, with $m\geq D$. 
These points are not treated as random, and the inference method we develop does not require any specific choice of $\{z_j\}_{j=1}^m$. However, to ensure that there is a reasonable variability in the values of $k(x_i,z_j)$, these points should be placed in the high density regions of $\p$. The simplest approach is to use a small held out portion of the data (with $m\ll n$ but $m\geq D$). Now, when we evaluate $\widehat{\mu_\theta}$ at these points, our modelling assumption from \eqref{eq:like1} on vector $\widehat{\mu_\theta}({\bf z})=\left[\widehat{\mu_\theta}(z_{1}),\ldots,\widehat{\mu_\theta}(z_{m})\right]$ can be written as
\begin{equation}
\widehat{\mu_\theta}({\bf z})|\mu_\theta\sim\mathcal{N}\left(\mu_\theta({\bf z}),\frac{\tau^{2}}{n}I_{m}\right).
\end{equation}
However, as $\widehat{\mu_\theta}(z_{j})=\frac{1}{n}\sum_{i=1}^n k_\theta(X_i,z_j)$ and all the terms $k_\theta(X_i,z_j)$ are independent given $\mu_\theta$, by Cram\'er's decomposition theorem, this modelling assumption is for the mapping $\phi_{{\bf z}}:\mathbb R^D\mapsto \mathbb R^m$, given by
$$\phi_{{\bf z}}(x):=\left[k_\theta(x,z_{1}),\ldots,k_\theta(x,z_{m})\right]\in\mathbb{R}^{m},$$
equivalent to:
\begin{equation}
\phi_{{\bf z}}(X_{i})|\mu_\theta\sim\mathcal{N}\left(\mu_\theta({\bf z}),\tau^{2}I_{m}\right).\label{eq:individual_normal}
\end{equation}

Applying the change of variable $x\mapsto \phi_{{\bf z}}(x)$ and using the generalization of the change-of-variables formula
to non-square Jacobian matrices as described in \citep{ben1999change},
we obtain a distribution for $x$ conditionally on  $\mu_\theta$ and $\theta$:
\begin{equation}
p(x|\mu_\theta,\theta)=p\left(\phi_{{\bf z}}(x)|\mu_\theta({\bf z})\right)\text{vol}\left[J_\theta(x)\right],
\label{eq:like2}
\end{equation}
where $J_\theta(x)=\left[\frac{\partial k_\theta(x,z_{i})}{\partial x^{(j)}}\right]_{ij}$
is an $m\times D$ matrix, and
\begin{eqnarray}
\text{vol}\left[J_\theta(x)\right] & = & \left(\det\left[J_\theta(x)^{\top}J_\theta(x)\right]\right)^{1/2}\nonumber\\
 & = & \left(\det\left[\sum_{l=1}^{m}\frac{\partial k_\theta(x,z_{l})}{\partial x^{(i)}}\frac{\partial k_\theta(x,z_{l})}{\partial x^{(j)}}\right]_{ij}\right)^{1/2}\nonumber\\
 &=: &\gamma_\theta(x)\;.\label{eq:jacobian2}
\end{eqnarray}
The notation $\gamma_\theta(x)$ highlights the dependence on both $\theta$ and $x$. An explicit calculation of $\gamma_\theta(x)$ 
   for squared exponential kernels is described in Section \ref{section:calculations}.

By the conditional independence of $\{\phi_{{\bf z}}(X_i)\}_{i=1}^n$ given $\mu_\theta$, we obtain the pseudolikelihood of all $n$ observations:
\begin{align}
\label{eq:pseudolikelihood}
&p(x_{1},\ldots,x_{n}|\mu_\theta,\theta)  =  \prod_{i=1}^{n} \mathcal{N}\left(\phi_{{\bf z}}(x_{i});\mu_\theta({\bf z}),\tau^{2}I_{m}\right)\gamma_\theta(x_i) \nonumber\\
 &\; =  \mathcal{N}\left(\phi_{{\bf z}}({\bf x});{\bf m}_\theta({\bf z}),\tau^{2}I_{mn}\right)\prod_{i=1}^{n}\gamma_\theta(x_i),
\end{align}
where 
\begin{equation*}
\phi_{{\bf z}}({\bf x})=\left[\phi_{{\bf z}}(x_{1})^{\top}\cdots\phi_{{\bf z}}(x_{n})^{\top}\right]^{\top}=\text{vec}\left\{K_{\theta,\bf zx }\right\}\in\mathbb{R}^{mn} 
\end{equation*}
and in the mean vector ${\bf m}_\theta({\bf z})=\left[\mu_\theta({\bf z})^{\top}\cdots\mu_\theta({\bf z})^{\top}\right]^{\top}$, $\mu_\theta({\bf z})$ repeats $n$ times. Under the prior \eqref{eq:prior}, this mean vector has
mean $\bf 0$ and covariance ${\bf 1}_{n}{\bf 1}_{n}^{\top}\otimes R_{\theta,{\bf zz}}$ where $R_{\theta,{\bf zz}}$ is the $m\times m$ matrix such that its $(i,j)$-th element is $r_\theta(z_i,z_j)$. Combining this prior and the pseudolikelihood in \eqref{eq:pseudolikelihood}, we have the marginal pseudolikelihood:
\begin{align}
    &p(x_1,\ldots,x_n|\theta)   =\int p(x_1,\ldots,x_n|\mu_\theta,\theta)p(\mu_\theta|\theta) d\mu_\theta \nonumber\\
                             & \;=\int \mathcal{N}\left(\phi_{{\bf z}}({\bf x});{\bf m}_\theta({\bf z}),\tau^{2}I_{mn}\right)\left[\prod_{i=1}^{n}\gamma_\theta(x_i)\right]p(\mu_\theta|\theta) d\mu_\theta                           \nonumber\\
                               & \;=\mathcal{N}\left(\phi_{{\bf z}}({\bf x});{\bf 0},{\bf 1}_{n}{\bf 1}_{n}^{\top}\otimes R_{\theta,{\bf zz}}+\tau^{2}I_{mn}\right)\prod_{i=1}^{n}\gamma_\theta(x_i). \label{eq:integrate}
\end{align}

While the marginal pseudolikelihood in Eq.~\eqref{eq:integrate} involves a computation of the likelihood for an $mn$-dimensional normal distribution, the Kronecker structure of the covariance matrix allows efficient computation as described in Appendix \ref{sec:Kronecker}. The complexity for calculating this likelihood is $\mathcal{O}(m^3+mn)$ (dominated by the inversion of $R_{\theta,{\bf zz}}+(\tau^2/n)I_m$).  The Jacobian term depends on the parametric form of $k_{\theta}$, but a typical cost
as shown in Section \ref{section:calculations} for the squared exponential kernel is $\mathcal{O}(nD^3+nmD^2)$. 
In this case, the computation of matrices $R_{\theta,{\bf zz}}$ and $\phi_{{\bf z}}({\bf x})=\text{vec}\left\{K_{\theta,\bf zx }\right\}$ is $\mathcal{O}(m^2D)$ and $\mathcal{O}(mnD)$ respectively.

Just as in GP modeling, the marginal pseudolikelihood can be maximized directly 
for maximum likelihood II (also known as empirical Bayes) estimation, in which we look for a single
best $\hat\theta$, or it can be used to construct 
an efficient MCMC sampler from the posterior of $\theta$. 

    \subsection{EXPLICIT CALCULATIONS FOR SQUARED EXPONENTIAL (RBF) KERNEL}
    \label{section:calculations}
Consider the isotropic squared exponential kernel with lengthscale matrix $\theta^2 I_D$  defined by 
\begin{equation}
     k_\theta(x,y) = \exp(-.5 (x-y)^{\top} \theta^{-2} I_D (x-y)).
\end{equation}
In this case, we can analytically calculate $r_{\theta}(x,y)$,
exact form is given in the Appendix in Section \ref{section:r-calculation}.

The partial derivatives of $k_\theta(x,y)$ with respect to $x^{(i)}$ for $i=1,\ldots D$  can be easily derived as
$$
\frac{\partial k_\theta(x,y)}{\partial x^{(i)}}= k_\theta(x,y) \frac{x^{(i)} - y^{(i)}}{\theta^2}
$$
and therefore the Jacobian from Eq.~\eqref{eq:jacobian2} is equal to
\begin{align}
 \gamma_\theta(x) &=  \left(\det\left[\sum_{l=1}^{m} k_\theta(x,z_l)^2 \;\frac{ (x^{(i)} - z_l^{(j)})^2}{\theta^4} \right]_{ij}\right)^{1/2}\;.
\end{align}
The computation of the matrix is $\mathcal{O}(mD^2)$ and the determinant is $\mathcal{O}(D^3)$. Since we must calculate $\gamma_\theta(x_i)$ for each $x_i$, the overall time complexity is $\mathcal{O}(nD^3+nmD^2)$.

\section{EXPERIMENTS}
\label{section:experiments}
We demonstrate our approach on two synthetic datasets and one example on real data, focusing on
two-sample testing with MMD and independence testing with HSIC. 
First, we use our Bayesian Kernel Embedding model and learn the kernel hyperparameters
with maximum likelihood II, optimizing the marginal likelihood. Second, we take a fully Bayesian
approach to inference and learning with our model. Finally, we apply the PC algorithm for causal structure discovery to a real dataset. The PC algorithm relies on a series of independence tests; we use HSIC with the lengthscales set with Bayesian Kernel Learning.
\begin{figure*}[ht!]
   \centering
    \includegraphics[width=.85\paperwidth]{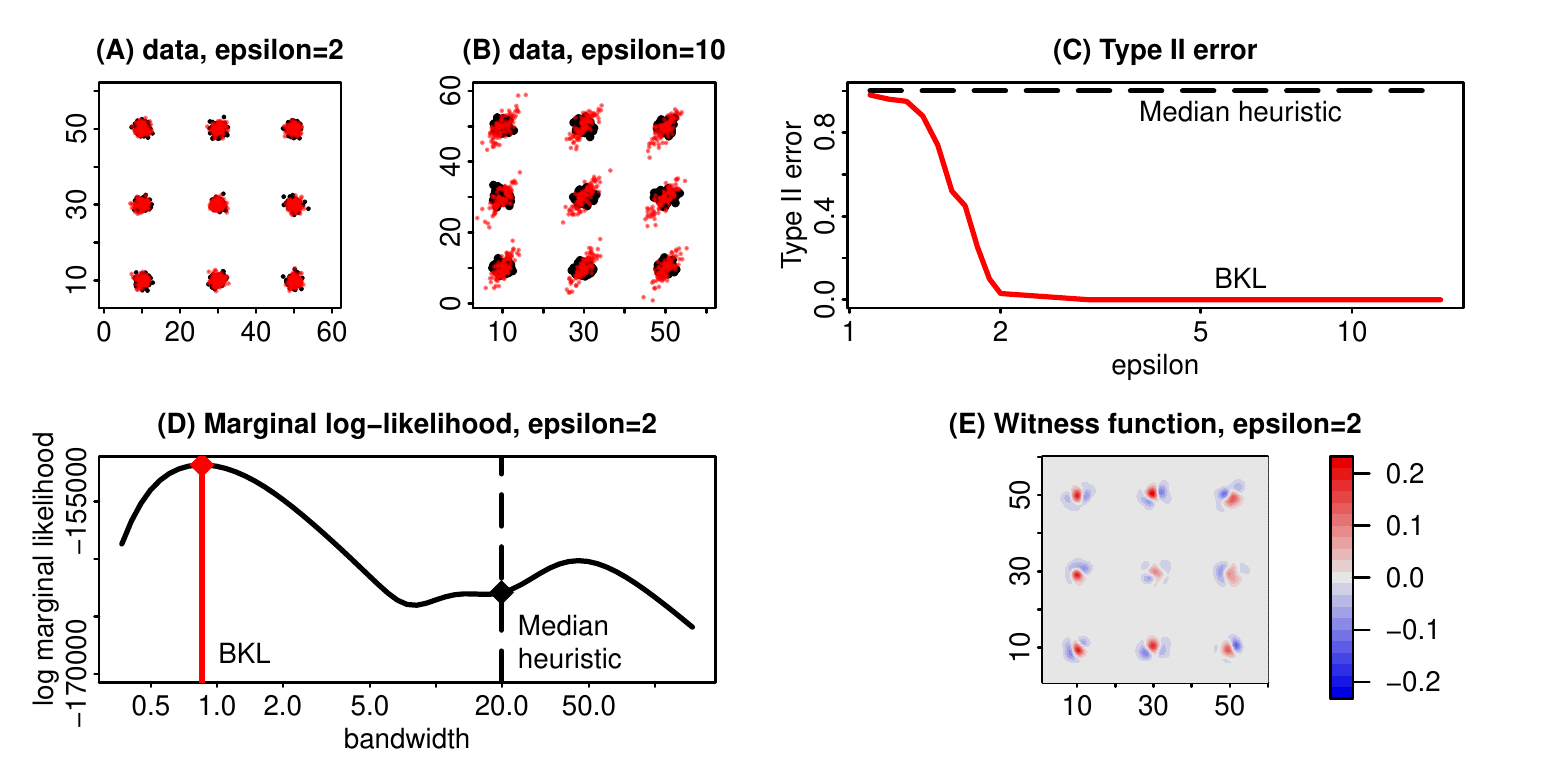}
    \caption{Two sample testing on a challenging simulated data set: comparing samples from a grid of isotropic Gaussians (black dots) to samples from a grid of non-isotropic Gaussians (red dots) with a ratio $\epsilon$ of largest to smallest covariance eigenvalues. Panels  (A) and (B) illustrate such samples for two values of $\epsilon$. (C) Type II error as a function of $\epsilon$ for significant level $\alpha=0.05$ following the median heuristic or the BKL approach to choose the lengthscale. (D) BKL marginal log-likelihood across a range of lengthscales. It is maximised for a lengthscale of 0.85 whereas the median heuristic suggests a value of 20.  (E) Witness function for the difficult case where $\epsilon=2$ using the BKL lengthscale. }
    \label{fig:MixtureGaussian}
\end{figure*}

Choosing lengthscales with the median heuristic is often a very bad idea. In the case of two sample testing,
\cite{gretton2012optimal} showed that MMD with the median heuristic failed to reject the null hypothesis when
    comparing samples from a grid of isotropic Gaussians to samples from a grid of non-isotropic Gaussians. We repeated this experiment by considering a distribution $\p$ of a mixture of bivariate Gaussians centered on a grid with diagonal covariance and unit variance and a distribution $\q$ of a mixture of bivariate Gaussians centered at the same locations but with rotated covariance matrices with a ratio $\epsilon$ of largest to smallest covariance eigenvalues. 
    
    As illustrated in Figures \ref{fig:MixtureGaussian}(A) and (B), for small values of $\epsilon$ both distributions are very similar whereas the distinction between $\p$ and $\q$ becomes more apparent as $\epsilon$ increases. For different values of $\epsilon$, we sample 100 observations from each mixture component, yielding 900 observations from $\p$ and 900 observations from $\q$ and then perform a two-sample test (${\bf H}_0: \p = \q$ vs. ${\bf H}_1: \p \neq \q$) using the MMD empirical estimate with an isotropic squared exponential kernel with one hyperparameter, the lengthscale. The type II error (i.e. probability that the test fails to reject the null hypothesis that  $\p=\q$ at $\alpha = 0.05$) is shown in Figure \ref{fig:MixtureGaussian}(C) for differently skewed covariances ($\epsilon$ from 0.5 to 15) when the median heuristic is chosen to select the kernel lengthscale or when using the Bayesian Kernel Learning.  In this example, the median heuristic picks a kernel with a large lengthscale, since the median distance between points is large. With this large lengthscale MMD always fails to reject at $\alpha = 0.05$ even for simple cases where $\epsilon$ is large. 
    When we use Bayesian Kernel Learning and optimize the marginal likelihood of Eq.~\eqref{eq:integrate} for $\tau^2 = 1$ (our results were not sensitive to the choice of this parameter, but in the fully Bayesian case below we show that we can learn it) we found the maximum marginal likelihood at a lengthscale of $0.85$. With this choice of lengthscale, MMD correctly rejects the null hypothesis at $\alpha = 0.05$ even for very hard situations when $\epsilon=2$. We observe that when $\epsilon$ is smaller than $2$, the type II error of MMD is very high for both choices of lengthscale, because
the two distributions $\p$ and $\q$ are so similar that the test always retains the null hypothesis.  
In Figure \ref{fig:MixtureGaussian}(D) we illustrate the BKL marginal likelihood across a range of lengthscales. Interestingly, there are multiple local optima and the median heuristic lies between the two main modes. The plot indicates that multiple scales may be of interest for this dataset, which makes sense given that the true data generating process is a mixture model.  This insight can be incorporated into the Bayesian Kernel Embedding framework by expanding our model, as discussed below. In Figure \ref{fig:MixtureGaussian}(E) we used the BKE posterior to estimate the witness function $\mu_{P,\theta}-\mu_{Q,\theta}$. 
This function is large in magnitude in the locations where the two distributions differ. For ease of visualization we do not try to
include posterior uncertainty intervals, but these are readily available from our model, and we show them for a 1-dimensional case below.

\begin{figure}[hb!]
   \centering
    \includegraphics[width=.8\textwidth]{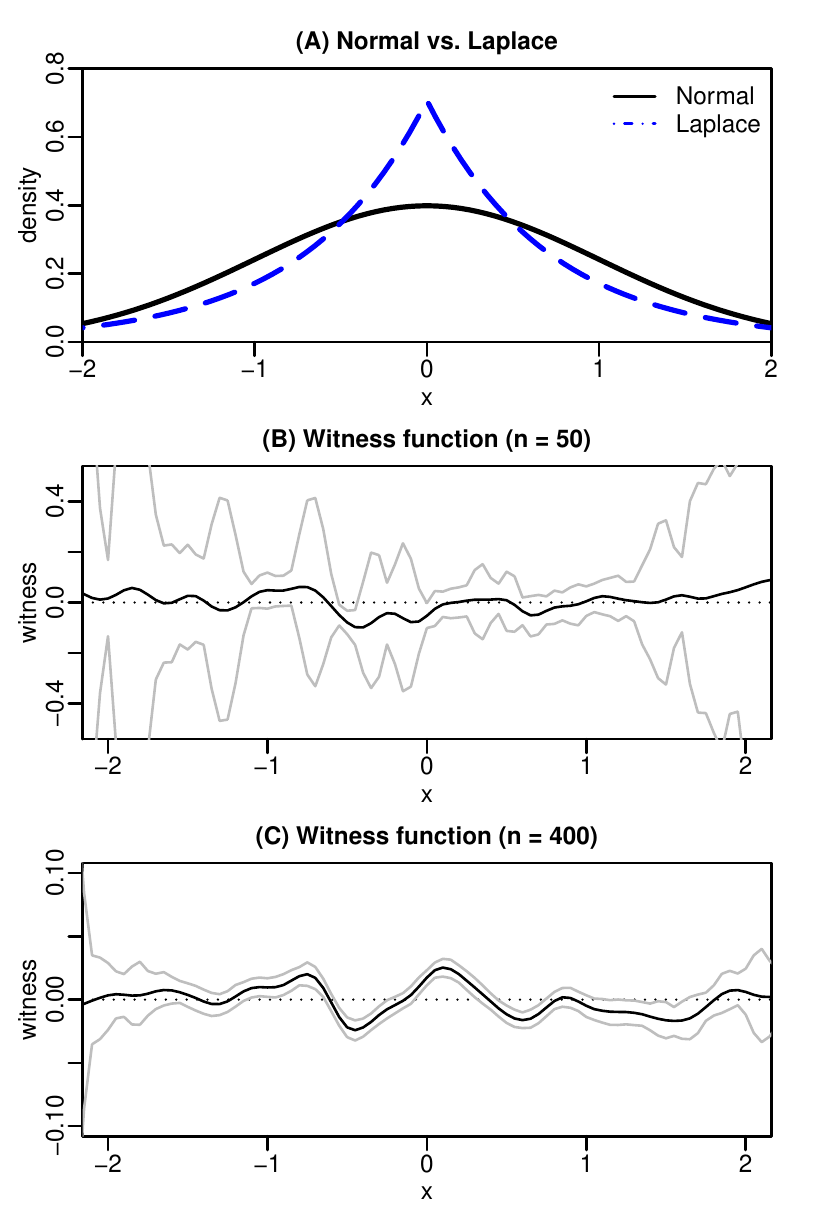}
    \caption{The true data generating process is shown in (A) where two samples of size $n$ are drawn from distributions with equal means and variances. We then fit our Bayesian Kernel Embedding model, with priors over the hyperparameters $\theta$ and $\tau^2$ to obtain a posterior over the witness function  for two-sampling testing. The witness function indicates the model's posterior estimates of where the two distributions differ (when the witness function is zero, it indicates no difference between the distributions). Posterior means and 80\% uncertainty intervals are shown. In (B) the small sample size means that the model does not effectively distinguish between samples from a normal and a Laplace distribution, while in (C) larger samples enable the model to find a clear difference, with much of the uncertainty envelope excluding 0.}
    \label{fig:bayesian-witness}
\end{figure}

\begin{figure}[hb!]
   \centering
    \includegraphics[width=.8\textwidth]{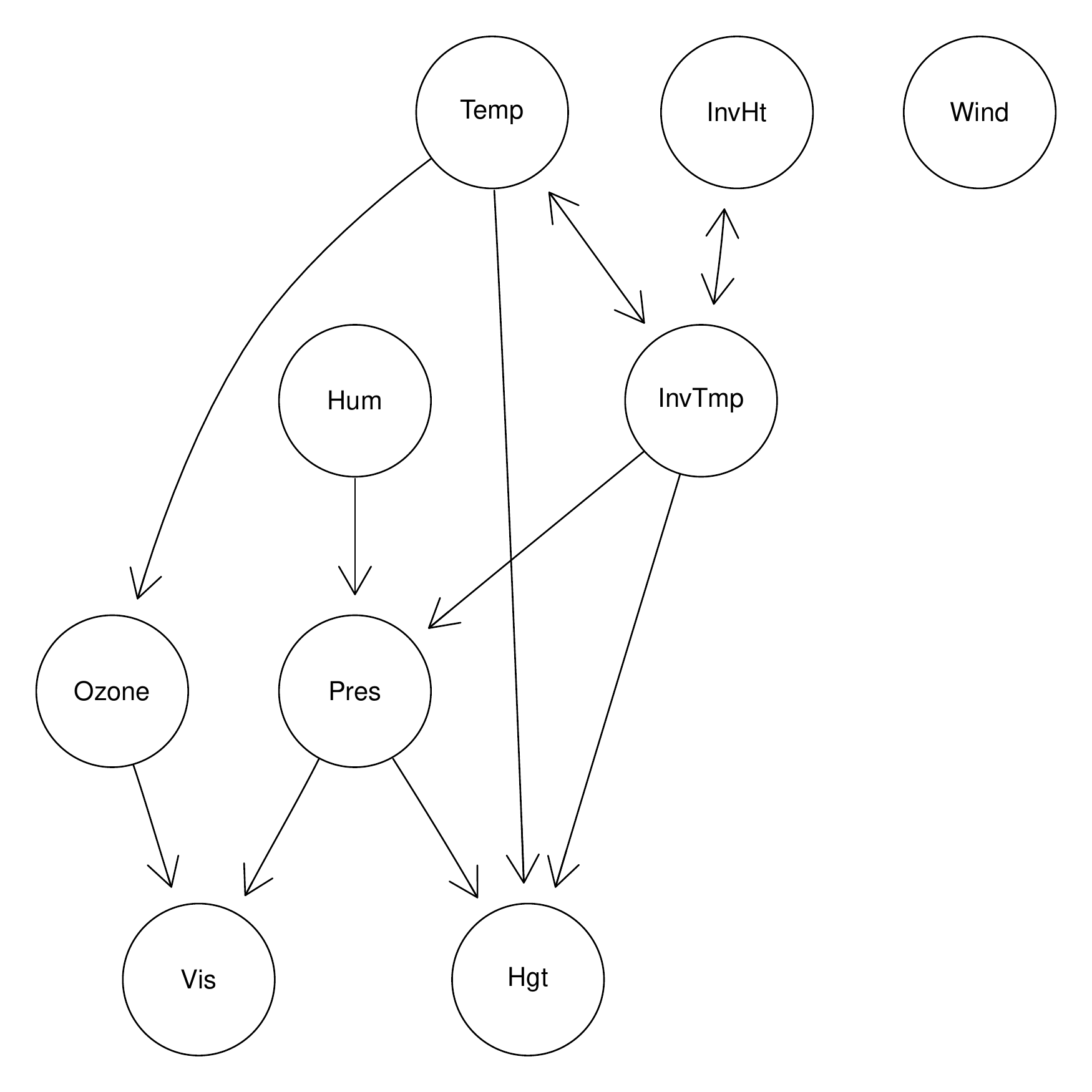}
    \caption{Graphical model representing an equivalence class of DAGs for the Ozone dataset from \protect{\cite{breiman1985estimating}}, learned using the PC algorithm following
    the approach in \cite{flaxman2015gaussian} with HSIC to test for independence. We used BKL to set hyperparameters of HSIC. Singly directed edges represent causal links, while bidirected edges represent edges that the algorithm failed to orient. The causal edge from temperature to ozone accords with scientific understanding, and was not present in the graphical model learned in \cite{flaxman2015gaussian} which employed the median heuristic.}
    \label{fig:ozone}
\end{figure}

Our model does not just provide a better way of choosing lengthscales. We can also use it in a fully Bayesian context,
where we place priors over the hyperparameters $\theta$ and $\tau^2$, and then integrate them out
to learn a posterior distribution over the mean embedding.
Switching to one dimension, we consider a distribution $\p = \mathcal{N}(0,1)$
and a distribution $\q = \mbox{Laplace}(0,\sqrt{.5})$. 
The densities are shown in Figure \ref{fig:bayesian-witness}(A).
Notice that the first two moments of these distributions are equal. 
To create a synthetic dataset we sampled $n$ observations from each distribution, and then combined them together into a sample of size $2n$, following the strategy in the previous experiment to learn a single lengthscale and kernel mean embedding for the combined dataset.  We ran a Hamiltonian Monte Carlo sampler (HMC) with NUTS (Stan source code is in the Appendix in Section \ref{section:stan:model}) for the Bayesian Kernel Embedding model with a squared exponential kernel, placing a $\mbox{Gamma}(1,1)$ prior on the lengthscale $\theta$
of the kernel and a $\mbox{Gamma}(1,1)$ prior on $\tau^2$. We ran 4 chains for 400 iterations, discarding 200 iterations as warmup, with the chains starting at different random initial values. Standard convergence and mixing diagnostics were good ($\hat R \approx 1$), so we considered the result to be 800 draws from the posterior distribution. Recall that for fixed hyperparameters
$\theta$ and $\tau^2$ we can obtain a posterior distribution over $\mu_{P,\theta}$ and $\mu_{Q,\theta}$. For each
of our 800 draws, we drew a sample from these two distributions and then calculated the witness function as the
difference, thus obtaining a random function drawn from the posterior distribution over $\mu_{P,\theta} - \mu_{Q,\theta}$
(where in practice we evaluate this function at a fine grid for plotting purposes).
We thus obtained the full posterior distribution over the witness function, integrating over the kernel hyperparameter.
We followed this procedure twice to create a dataset with $n = 50$ and a dataset with $n = 400$.
In Figure \ref{fig:bayesian-witness}(B) we see that the witness function for the small dataset is not able to distinguish
between the distributions as it rarely excludes 0. (Note that our model has the function 0 as its prior, which corresponds to the null hypothesis that the two distributions are equal. This could easily be changed to incorporate any relevant prior information.). As shown in Figure \ref{fig:bayesian-witness}(C), with more data the witness function is able to distinguish between
the two distributions, mostly excluding 0.
Finally, we consider the ozone dataset analyzed in \cite{breiman1985estimating}, consisting of daily measurements of ozone concentration and eight related meteorological variables. Following the approach
in \cite{flaxman2015gaussian}, we first pre-whiten the data to control for underlying temporal autocorrelation,
then we use a combination of Gaussian process regression followed by HSIC to test for conditional independence. Each time we run HSIC, we set the kernel hyperparameters using Bayesian Kernel Learning. The graphical model that we learn is shown in Figure \ref{fig:ozone}. The directed edge from the temperature variable to ozone is encouraging, as
higher temperatures favor ozone formation through a variety of chemical processes which are not represented by variables in this dataset \citep{bloomer2009observed,sillman1999relation}. Note that this edge was not present in the graphical model in \cite{flaxman2015gaussian} in which the median heuristic was used. 

\section{DISCUSSION}
\label{section:discussion}
We developed a framework for Bayesian learning of kernel embeddings of probability measures. It is primarily designed for unsupervised settings, and in particular for kernel-based hypothesis testing.
In these settings, one relies critically on a good choice of kernel and our framework yields a new method, termed Bayesian Kernel Learning, to inform this choice.
We only explored learning the lengthscale of the squared exponential kernel, but our method 
extends to the case of richer kernels with more hyperparameters. We conceive of Bayesian Kernel Learning as a drop-in replacement for
selecting the kernel hyperparameters in settings where cross-validation is unavailable.
A sampling-based Bayesian approach is also demonstrated, enabling integration over kernel hyperparameters, and e.g., 
obtaining the full posterior distribution over the witness function in two-sample testing.

While our method is designed for unsupervised settings, there are various
reasons it might be helpful in supervised settings or in applied Bayesian modelling more generally.
With the rise of large-scale kernel methods, it has become possible to apply, e.g.~SVMs or GPs
to very large datasets.  But even with efficient methods, it can be very costly to run cross-validation over a large space of
hyperparameters.  In practice, when, e.g.~large scale approximations based on random Fourier features \citep{Rahimi07randomfeatures} are used, we have
not seen much attention paid to kernel learning -- the features are often just
one part of a complicated pipeline, so again the median heuristic is often
employed. For these reasons, we think that the developed method for Bayesian Kernel Learning would be a judicious alternative.
Moreover, it would be straightforward to develop scalable approximate versions of
Bayesian Kernel Learning itself. 

\section{Acknowledgments}
SRF was supported by the ERC (FP7/617071) and EPSRC (EP/K009362/1).
Thanks to Wittawat Jitkrittum, Krikamol Muandet, Sayan Mukherjee, Jonas Peters, Aaditya Ramdas,
Alex Smola, and Yee Whye Teh for helpful discussions.

\clearpage
{

\bibliography{biblio}
}

\onecolumn
\appendix
\section{Some derivations for Bayesian Kernel Embedding}

\subsection{Notation}
Consider a dataset $x_1, \ldots, x_n\in  \R^D$ and suppose that there exists some unknown probability distribution
$\p$ for which the $x_i$ are i.i.d.:
\begin{equation}
    x_i \sim \p\;.
\end{equation}
Denote by $\mu_\theta$  the RKHS mean embedding element for a given kernel $k_\theta(\cdot,\cdot)$ with hyperparameter $\theta\in\R^Q$ and by $\widehat{\mu_\theta}(\cdot)$ the empirical mean embedding 
\begin{equation}
   \widehat{\mu_\theta}(\cdot) := \frac{1}{n} \sum_{i=1}^n k_\theta(x_i,\cdot)\;.
   \end{equation}

We posit as our model that $\mu_\theta$ has a GP prior with covariance $r_\theta$,
where $$r_{\theta}(x,y) = \int k_\theta(x,u)k_\theta(u,y) \nu(du)\;,$$
where $\nu$ is a finite measure on $\R^D$ thus ensuring that $\mu_\theta \in \mathcal{H}_{k_{\theta}}$ when drawn from the prior 
\begin{equation}
\mu_\theta | \theta \sim \mathcal{GP}(0, r_\theta(\cdot,\cdot))\;.
\label{eq:f-prior}
\end{equation}
 In addition, we model the link between the population mean embedding and the empirical mean embedding
functions at a given location $x$ as follows\begin{equation}
p(\widehat{\mu_\theta}(x)|\mu_\theta(x)) = \mathcal{N}(\widehat{\mu_\theta}(x); \mu_\theta(x), \tau^2/n)
\label{eq:like1_app}
\end{equation}
where $\tau^2$ is another hyperparameter.

\subsection{Priors over RKHS}
\label{section:nuclear-dominance}
The results in this section have appeared in the literature before, but as they are not well known or 
collected in one place, we have included them for completeness. A similar discussion appears in \cite{pillai2007characterizing}, but without the construction of explicit GP priors over the RKHSs which we provide below.

It is well known that the sample paths of a GP with kernel $k$ are almost surely outside RKHS $\rkhs$, the result known as Kallianpur's 0-1 law \citep{kallianpur, wahba1990}. It is easiest to demonstrate this by considering a Mercer's expansion \citet[Section 4.3]{rasmussen2006gaussian} of kernel $k$ given by
\begin{equation}
 k(x,x')=\sum_{i=1}^\infty \lambda_i e_i(x)e_i(x'),
\end{equation}
for the eigenvalue-eigenfunction pairs $\{(\lambda_i,e_i)\}_{i=1}^n$.
Then, a representation of $f\sim \mathcal{GP}(0,k)$ is given by $f=\sum_{i=1}^\infty \sqrt{\lambda_i}Z_ie_i$, where $\{Z_i\}_{i=1}^\infty$ are independent and identically distributed standard normal random variables. However, 
\begin{equation}
\| f \|^2_{\mathcal H_k}=\sum_{i=1}^\infty \frac{\lambda_i Z_i^2}{\lambda_i}=\sum_{i=1}^\infty Z_i^2 = \infty, \quad a.s.
\end{equation}
so $f\not\in\rkhs$ almost surely. This issue is often sidelined in the literature, cf. e.g. \cite[Section 6.1]{rasmussen2006gaussian} -- in GP regression, it is not necessary to ensure that the prior on the regression function is supported on $\rkhs$ (the posterior mean will still lie in $\rkhs$, however). However, since the object of our interest, kernel embedding, is by construction an element of $\rkhs$ - we opt for an approach where the prior is indeed specified over the correct space.  Fortunately, it is straightforward to construct a kernel $r$ such that the realizations from a GP with kernel $r$ are almost surely inside RKHS $\rkhs$. For this, we will need notions of dominance and nuclear dominance for kernel functions.
\begin{definition}
 Kernel $k$ is said to dominate kernel $r$ (written $k\succ r$) if $\mathcal{H}_r \subseteq \mathcal{H}_k$.
\end{definition}
\citet[Theorem 1.1]{Lukic:2001} characterise dominance $k\succ r$ via the existence of a certain positive, continuous and self-adjoint operator $L:\rkhs\to\rkhs$ for which
\begin{equation}
 r(x,x')=\langle L[k(\cdot, x)],k(\cdot, x') \rangle_{\rkhs},\qquad\forall x,x'\in\domX.
\end{equation}
When $L$ is also a trace class operator, dominance is termed \emph{nuclear}, and denoted $k\succ\succ r$.
The following theorem from \citet[Theorem 7.2]{Lukic:2001} then fully characterises kernels that lead to valid GP priors over RKHS $\rkhs$.
\begin{theorem}
 Let $\rkhs$ be separable and let $m\in\rkhs$. Then $\mathcal{GP}(0, r(\cdot,\cdot))$ has trajectories in $\rkhs$ with probability 1 if and only if $k\succ\succ r$.
\end{theorem}
Thus, we just need to specify a trace-class, positive, continuous and self-adjoint operator $L:\rkhs\to\rkhs$ and compute $\langle L[k(\cdot, x)],k(\cdot, x') \rangle_{\rkhs}$. A convenient choice for a given bounded continuous kernel $k$ can be defined as follows. Take the convolution operator $S_k:L^2(\mathcal X; \nu)\to \rkhs$ with respect to a finite measure $\nu$, defined as
\begin{equation}
 [S_k f](x)=\int f(u)k(x,u)\nu(du).
\end{equation}
It is well known that the adjoint of $S_k$ is the inclusion of $\rkhs$ into $L^2$ \citep[Section 4.3]{steinwart2008}. Thus, we let $L=S_kS_k^*$, which is the (uncentred) covariance operator $L=\int k(\cdot,u)\otimes k(\cdot,u)\nu(du)$ of $\nu$. As a covariance operator, $L$ is then positive, continuous and self-adjoint. It is also trace-class in most cases of interest -- and in particular whenever $\int k(u,u)\nu(du)<\infty$ \citep[Theorem 4.27]{steinwart2008}, and thus for every stationary kernel provided that $\nu$ is a finite measure. This leads to
\begin{eqnarray*}
 r(x,x')&=&\langle S_kS_k^*[k(\cdot, x)],k(\cdot, x') \rangle_{\rkhs}\\
 {}&=&\langle S_k^*[k(\cdot, x)],S_k^*k(\cdot, x') \rangle_{L^2(\mathcal X; \nu)}\\
 {}&=&\int k(x,u)k(u,x')\nu(du),
\end{eqnarray*}
so $r$ can be simply computed as a convolution of $k$ with itself, and we can use $\mathcal{GP}(0, r(\cdot,\cdot))$ as a prior over $\rkhs$.

\subsection{Covariance function $r_\theta$}
\label{section:r-calculation}
In this subsection, we derive the covariance function $r_\theta$ for squared exponential kernels. Consider a squared exponential kernel on $\mathcal X=\mathbb R^D$ with full covariance matrix $\Sigma_\theta$ defined by 
\begin{equation}
     k_\theta(x,y) = \exp\left(-\frac{1}{2}(x-y)^T\Sigma_\theta^{-1}(x-y)\right),\quad x, y\in \R^D.
\end{equation}
While we have required in \ref{section:nuclear-dominance} that $\nu$ is a finite measure for the covariance operator to be trace class when working with stationary kernels, let us for simplicity first consider the instructive case when $\nu$ is the Lebesgue measure. Then, we have
\begin{align*}
r_\theta(x,y)&=\int k_\theta(x,u) k_\theta(u,y) du\\
&=\int  \exp\left(-\frac{1}{2}\left((x-u)^T\Sigma_\theta^{-1}(x-u)+(y-u)^T\Sigma_\theta^{-1}(y-u)\right)\right)du
\end{align*}
Note that
$$(x-u)^T\Sigma_\theta^{-1}(x-u)+(y-u)^T\Sigma_\theta^{-1}(y-u)=2\left(u-\frac{x+y}{2}\right)^T\Sigma_\theta^{-1}\left(u-\frac{x+y}{2}\right)+\frac{1}{2} (x-y)^T\Sigma_\theta^{-1}(x-y)\;.$$
Then
\begin{align*}
r_\theta(x,y)&=\exp\left(-\frac{1}{2} (x-y)^T(2\Sigma_\theta)^{-1}(x-y)\right)\int \exp\left(-\frac{1}{2}\left(u-\frac{x+y}{2}\right)^T\left(\frac{1}{2}\Sigma_\theta\right)^{-1}\left(u-\frac{x+y}{2}\right)\right)du\\
&=\exp\left(-\frac{1}{2} (x-y)^T(2\Sigma_\theta)^{-1}(x-y)\right) \; \times (2\pi)^{D/2}|\Sigma_\theta/2|^{1/2}\\
&=\pi^{D/2}\;|\Sigma_\theta|^{1/2}\;\exp\left(-\frac{1}{2}(x-y)^T(2\Sigma_\theta)^{-1}(x-y)\right)\;.
\end{align*}
Thus $r_\theta$ is proportional to another squared exponential kernel with covariance $2\Sigma_\theta$.
For the special case where the covariance matrix $\Sigma_\theta$ is diagonal -- let $\Sigma_\theta=\theta I_D$ and $\theta=(\theta^{(1)},\dots,\theta^{(D)})^T$-- we have
\begin{equation}
r_\theta(x,y)=\pi^{D/2}\;\left(\prod_{d=1}^D\theta^{(d)}\right)^{1/2}\exp\left(-\frac{1}{2}(x-y)^T(2\theta I_D)^{-1}(x-y)\right).
\end{equation}

Now, take $\nu(du)=\exp\left(-\frac{\Vert u \Vert_2^2}{2\eta^2}\right)du$, i.e., $\nu$ is a finite measure and is proportional to a Gaussian measure on $\mathbb R^d$. In that case, we have 
\begin{align*}
r_\theta(x,y)&=\int k_\theta(x,u) k_\theta(u,y) \nu(du)\\
&=\int  \exp\left(-\frac{1}{2}\underset{{\mathsf A}}{\underbrace{\left((x-u)^T\Sigma_\theta^{-1}(x-u)+(y-u)^T\Sigma_\theta^{-1}(y-u)+\eta^{-2}u^\top u\right)}}\right)du.
\end{align*}
From standard Gaussian integration rules, it follows that 
\begin{align*}
 \mathsf{A}=\frac{1}{2} (x-y)^T\Sigma_\theta^{-1}(x-y)+(u-{\mathsf m})^\top {\sfS}^{-1}(u-{\mathsf m})+\left(\frac{x+y}{2}\right)^\top\left(\frac{1}{2}\Sigma_\theta+\eta^2 I_D\right)^{-1}\left(\frac{x+y}{2}\right)
\end{align*}
where $\mathsf{m}=\sfS^{-1}\Sigma_\theta^{-1}\left(x+y\right)$ and $\sfS=(2\Sigma_\theta^{-1}+\eta^{-2}I_D)^{-1}$.
Therefore
\begin{align*}
r_\theta(x,y)&=(2\pi)^{D/2}|\sfS|^{1/2}\exp\left(-\frac{1}{2} (x-y)^T(2\Sigma_\theta)^{-1}(x-y) -\frac{1}{2}\left(\frac{x+y}{2}\right)^\top\left(\frac{1}{2}\Sigma_\theta+\eta^2 I_D\right)^{-1}\left(\frac{x+y}{2}\right)
 \right)\\
 &=(2\pi)^{D/2}\left|2\Sigma_\theta^{-1}+\eta^{-2}I_D\right|^{-1/2}\exp\left(-\frac{1}{2} (x-y)^T(2\Sigma_\theta)^{-1}(x-y)\right)\\
 &\qquad\qquad\qquad\qquad\qquad\qquad\qquad\times\exp\left( -\frac{1}{2}\left(\frac{x+y}{2}\right)^\top\left(\frac{1}{2}\Sigma_\theta+\eta^2 I_D\right)^{-1}\left(\frac{x+y}{2}\right) \right)\;.
\end{align*}
Thus, we see that $r_\theta$ has a nonstationary component that penalises the norm of $\left(\frac{x+y}{2}\right)$. This is reminiscent of the well known locally stationary covariance functions \citep{silverman1957}. However, for large values of $\eta$, the nonstationary component becomes negligible and $r_\theta$ reverts to being proportional to a standard squared exponential kernel with covariance $2\Sigma_\theta$, just like in the case of Lebesgue measure. We note that any choice of $\eta>0$ gives a valid prior over $\mathcal H_k$. Treating $\eta$ as another hyperparameter to be learned would be an interesting direction for future research.

\subsection{Fast computation of the marginal pseudolikelihood}
\label{sec:Kronecker}
The marginal pseudolikelihood in Eq.~\eqref{eq:integrate} requires computation of the likelihood for an $mn$-dimensional normal distribution
$$\mathcal{N}\left(\text{vec}\left\{K_{\theta,\bf zx }\right\};{\bf 0},\;{\bf 1}_{n}{\bf 1}_{n}^{\top}\otimes R_{\theta,{\bf zz}}+\tau^{2}I_{mn}\right).$$
However, the Kronecker product structure in the covariance matrix $C={\bf 1}_{n}{\bf 1}_{n}^{\top}\otimes R_{\theta,{\bf zz}}+\tau^{2}I_{mn}$ allows efficient computation. We denote with $R_{\theta,{\bf zz}}=Q\Lambda Q^\top$ the eigendecomposition of the matrix $R_{\theta,{\bf zz}}$ with $\Lambda=\text{diag}\left[\lambda_1,\ldots,\lambda_m\right]$. Note that ${\bf 1}_{n}{\bf 1}_{n}^\top$ is a rank-one matrix with the eigenvalue equal to $n$. Therefore $C$ has top $m$ eigenvalues equal to $n\lambda_i+\tau^2$, $i=1,\ldots,m$, and the remaining $n(m-1)$ all equal to $\tau^2$. Thus, the log-determinant is simply
\begin{equation}
 \log\text{det}C=\sum_{i=1}^m \log(n\lambda_i+\tau^2)+m(n-1)\log\tau^2=\log\text{det}\left[R_{\theta,{\bf zz}}+(\tau^{2}/n)I_{m}\right]+m\log n+m(n-1)\log\tau^{2}.
\end{equation}

Further, we need to compute $\text{vec}\left\{K_{\theta,\bf zx }\right\}^\top C^{-1}\text{vec}\left\{K_{\theta,\bf zx }\right\}$. By completing $b_1=n^{-1/2}{\bf 1}_n$ to an orthonormal basis $\{b_1,\ldots,b_n\}$ of $\mathbb R^n$ and forming the corresponding matrix $B=[b_1 \cdots b_n]$, and denoting by $\mathsf{n}$ an $n\times n$ matrix with $\mathsf{n}_{11}=n$ and $\mathsf{n}_{ij}=0$ elsewhere, we have that 
\begin{equation}
 C^{-1}=(B\otimes Q)(\mathsf{n}\otimes\Lambda+\tau^2 I_{nm})^{-1}(B\otimes Q)^\top.
\end{equation}
We now simply need to apply Kronecker identity $(B^\top\otimes Q^\top)\text{vec}\left\{K_{\theta,\bf zx }\right\}=\text{vec}\left\{Q^\top K_{\theta,\bf zx }B\right\}$, to obtain
\begin{align}
 &\text{vec}\left\{K_{\theta,\bf zx }\right\}^\top C^{-1}\text{vec}\left\{K_{\theta,\bf zx }\right\}=\text{vec}\left\{Q^\top K_{\theta,\bf zx }B\right\}^\top (\mathsf{n}\otimes\Lambda+\tau^2 I_{nm})^{-1}\text{vec}\left\{Q^\top K_{\theta,\bf zx }B\right\}\nonumber\\
 &\quad = \sum_{j=1}^m \frac{n^{-1}\left[Q^\top K_{\theta,\bf zx }{\bf 1}_n\right]_j^2}{n\lambda_j+\tau^2}+\frac{1}{\tau^2}\sum_{i=2}^n\sum_{j=1}^m \left[Q^\top K_{\theta,\bf zx }b_i\right]_j^2.
\end{align}

For the first term, we have
\begin{align}
 &\sum_{j=1}^m \frac{n^{-1}\left[Q^\top K_{\theta,\bf zx }{\bf 1}_n\right]_j^2}{n\lambda_j+\tau^2}=\sum_{j=1}^m\frac{\left[Q^\top\hat\mu({\bf z})\right]_j^2}{\lambda_j+\tau^2/n}=\sum_{j=1}^m\frac{\text{Tr}\left[\hat\mu({\bf z})\hat\mu({\bf z})^\top q_jq_j^\top\right]}{\lambda_j+\tau^2/n}\nonumber\\
 &\qquad = \hat\mu({\bf z})^\top \left(R_{\theta,{\bf zz}}+(\tau^2/n)I_m\right)^{-1}\hat\mu({\bf z}).
\end{align}

And for the second term:
\begin{eqnarray}
\frac{1}{\tau^{2}}\sum_{i=2}^{n}\sum_{j=1}^{m}\left[Q^{\top}K_{\theta,{\bf zx}}b_{i}\right]_{j}^{2}	&=&	\frac{1}{\tau^{2}}\sum_{j=1}^{m}\sum_{i=2}^{n}\left[q_{j}^{\top}K_{\theta,{\bf zx}}b_{i}\right]^{2}\nonumber\\
	& = &	\frac{1}{\tau^{2}}\sum_{j=1}^{m}\left\{ \left\Vert K_{\theta,{\bf xz}}q_{j}\right\Vert ^{2}-n\left(q_{j}^{\top}\hat{\mu}({\bf z})\right)^{2}\right\} \nonumber\\
	& = &	\frac{1}{\tau^{2}}\left\Vert K_{\theta,{\bf xz}}\right\Vert _{F}^{2}-\frac{n}{\tau^{2}}\left\Vert \hat{\mu}\left({\bf z}\right)\right\Vert ^{2}.
\end{eqnarray}

Altogether, the log-likehood is given by
\begin{eqnarray}
\log\left\{\mathcal{N}\left(\text{vec}\left\{ K_{\theta,{\bf zx}}\right\} ;{\bf 0},\;{\bf 1}_{n}{\bf 1}_{n}^{\top}\otimes R_{\theta,{\bf zz}}+\tau^{2}I_{mn}\right)\right\} &=& -\frac{1}{2}\Biggl\{\log\text{det}\left[R_{\theta,{\bf zz}}+(\tau^{2}/n)I_{m}\right]\\
 &&\quad\quad +\,\hat{\mu}({\bf z})^{\top}\left(R_{\theta,{\bf zz}}+(\tau^{2}/n)I_{m}\right)^{-1}\hat{\mu}({\bf z})\nonumber\\
 &&\quad\quad\quad +\,\frac{1}{\tau^{2}}\left\Vert K_{\theta,{\bf xz}}\right\Vert _{F}^{2}-\frac{n}{\tau^{2}}\left\Vert \hat{\mu}\left({\bf z}\right)\right\Vert ^{2}\nonumber\\
 &&\quad\quad\quad\quad+\,m\log n+m(n-1)\log\tau^{2}+mn\log(2\pi)\Biggr\}\nonumber.
\end{eqnarray}

\section{Source for Stan model}
\label{section:stan:model}
\lstset{
    keywordstyle=\color{blue}
  , basicstyle=\ttfamily\small
  , commentstyle={}
  , columns=flexible
  , showstringspaces=false
  }

\begin{lstlisting}
functions {
  // phi should be m x n
  real kron_multi_normal(matrix K,matrix R,matrix Q1,vector e1,int m,int n,real sigma2) {
    vector[m*n] e;
    matrix[m,m] Q2;
    vector[m] e2;
    vector[m] ones;
    vector[m*n] mv2;
    real mvp;
    real logdet;
    Q2 <-  eigenvectors_sym(R);
    e2 <- eigenvalues_sym(R);
    for(j in 1:m) {
      ones[j] <- 1;
      for(i in 1:n)
        e[(j-1)*n + i] <- 1/(e1[i] * e2[j] + sigma2);
    }
    mv2 <- to_vector((transpose(Q2) * transpose(K)) * Q1);
    mvp <- sum(mv2 .* e .* mv2);
    logdet <- sum(log(e2 .* (ones * n) + ones * sigma2)) + m * (n-1) * log(sigma2);
 
    return( - .5 * logdet - .5 * mvp);
  }
}

data {
  int<lower=1> n;
  int<lower=1> m;
  vector[n] x;
  vector[m] u;
}

transformed data {
  matrix[n,m] xu_dist2;
  matrix[m,m] u_dist2;
  matrix[n,n] ones;
  vector[n] zeros;
  matrix[n,n] Q1;
  vector[n] e1;
  
  for (i in 1:n) {
    zeros[i] <- 0;
    e1[i] <- 0;
    for (j in 1:n) 
      ones[i,j] <- 1;    
    for(j in 1:m)
      xu_dist2[i, j] <- square(x[i] - u[j]);
  }
  for(i in 1:m) {
    for(j in 1:m)
      u_dist2[i,j] <- square(u[i] - u[j]);
  }
  e1[1] <- n;
  Q1 <- eigenvectors_sym(ones);
}

parameters {
  real<lower=0> lengthscale;
  real<lower=0> sigma2;
}
transformed parameters {
  matrix[m,m] R;                              
  matrix[n,m] J;                              
  matrix[n,m] K;
  
//  R <- lengthscale * sqrt(pi()) * 
  R <- exp(- u_dist2/(4*lengthscale^2));  
  K <- exp(- xu_dist2/(2*lengthscale^2)); 
  J <- K .* K .* xu_dist2 / lengthscale^4;
}

model {
  for(i in 1:n) // Jacobian
    increment_log_prob(log(.5 * sum(J[i])));

  increment_log_prob(kron_multi_normal(K, R, Q1, e1, m, n, sigma2));
  lengthscale ~ gamma(1,1);
  sigma2 ~ gamma(1,1);
}
\end{lstlisting}

\end{document}